\documentclass[]{fairmeta}

\usepackage[utf8]{inputenc} 
\usepackage[T1]{fontenc}    
\usepackage{graphicx}
\usepackage{amsmath}
\usepackage{amssymb}
\usepackage{fancyhdr} 
\usepackage{url}            
\usepackage{booktabs}       
\usepackage{amsfonts}       
\usepackage{nicefrac}       
\usepackage{microtype}      
\usepackage{xcolor}         
\usepackage{multirow}
\usepackage{comment}
\usepackage{colortbl}
\usepackage{tocloft}
\usepackage[ruled,vlined]{algorithm2e}
\usepackage{wrapfig}
\usepackage{tabularx}
\usepackage{textcomp}
\usepackage{stfloats}
\usepackage{verbatim}
\usepackage{titlesec}
\usepackage{adjustbox}
\usepackage{pifont}
\usepackage{tikz}
\usepackage{color}
\usepackage{subcaption}

\RequirePackage{xspace}
\makeatletter
\DeclareRobustCommand\onedot{\futurelet\@let@token\@onedot}
\def\@onedot{\ifx\@let@token.\else.\null\fi\xspace}

\makeatother

\definecolor{lightblue}{rgb}{0.66, 0.85, 0.95}
\definecolor{blue}{RGB}{0, 0, 255}
\hypersetup{
    colorlinks=true,
    linkcolor=red,
    citecolor=blue,
}
\definecolor{c2}{HTML}{FBD9BD}
\definecolor{c3}{HTML}{fe793d}
\definecolor{c4}{HTML}{eedeb0}
\definecolor{rouse}{rgb}{0.981,0.961,0.941}
\usepackage[most]{tcolorbox}
\tcbset{colback=blue!5!white, colframe=blue!20!white, boxrule=0pt, arc=2mm, left=1mm, right=1mm, top=1mm, bottom=1mm}

\definecolor{adptorange}{RGB}{248, 205, 172}
\definecolor{cmpblue}{RGB}{189, 215, 238}
\definecolor{cmpblue}{RGB}{189, 215, 238}

\definecolor{our_red}{RGB}{232,157,160}
\definecolor{our_blue}{RGB}{136,206,230}
\definecolor{our_orange}{RGB}{246,200,168}
\definecolor{our_green}{RGB}{178,211,164}

\definecolor{attn_code0}{RGB}{247,215,200}
\definecolor{attn_code1}{RGB}{238,169,139}
\definecolor{mlp_code0}{RGB}{204,201,221}
\definecolor{mlp_code1}{RGB}{102,95,153}

\definecolor{token_blue}{RGB}{84, 120, 140}

\usepackage{pifont}       
\usepackage{bbding}       
\usepackage{fontawesome}
\usepackage{xspace}

\usepackage{float}

\usepackage{makecell} 

\newlength\savewidth

\newcolumntype{x}[1]{>{\centering\arraybackslash}p{#1pt}}
\newcolumntype{y}[1]{>{\raggedright\arraybackslash}p{#1pt}}
\newcolumntype{z}[1]{>{\raggedleft\arraybackslash}p{#1pt}}

\renewcommand{\paragraph}[1]{\vspace{1mm}\noindent\textbf{#1}}

\usepackage{colortbl}
\usepackage{xcolor}
\usepackage{wrapfig}

\renewcommand{\paragraph}[1]{\vspace{1.25mm}\noindent\textbf{#1}}

\usepackage{listings}

\definecolor{codeblue}{rgb}{0.25, 0.5, 0.5}
\definecolor{codekw}{rgb}{0.35, 0.35, 0.75}
\lstdefinestyle{Pytorch}{
    language = Python,
    backgroundcolor = \color{white},
    basicstyle = \fontsize{9pt}{8pt}\selectfont\ttfamily\bfseries,
    columns = fullflexible,
    aboveskip=1pt,
    belowskip=1pt,
    breaklines = true,
    captionpos = b,
    commentstyle = \color{codeblue},
    keywordstyle = \color{codekw},
}

\definecolor{green}{HTML}{009000}
\definecolor{red}{HTML}{ea4335}
\definecolor{mygold}{HTML}{feecb3}

\title{How Far Are Surgeons from Surgical World Models?\\
\large A Pilot Study on Zero-shot Surgical Video Generation with Expert Assessment}
\author[\dagger 1]{Zhen Chen}
\author[2]{Qing Xu}
\author[3]{Jinlin Wu}
\author[\dagger 4]{Biao Yang} 
\author[\dagger 5]{Yuhao Zhai}
\author[4]{Geng Guo} 
\author[5]{Jing Zhang} 
\author[5]{Yinlu Ding} 
\author[6]{Nassir Navab}
\author[7]{Jiebo Luo}

\affiliation[1]{Yale University}
\affiliation[2]{University of Nottingham}
\affiliation[3]{Institute of Automation, Chinese Academy of Sciences}
\affiliation[4]{Department of Neurosurgery, The First Hospital, Shanxi Medical University}
\affiliation[5]{Department of Gastrointestinal Surgery, The Second Qilu Hospital, Shandong University}
\affiliation[6]{Technical University of Munich}
\affiliation[7]{University of Rochester}

\contribution[\dagger]{Corresponding author}

\abstract{
Foundation models in video generation are demonstrating remarkable capabilities as potential \textbf{world models} for simulating the physical world. However, their application in high-stakes domains like surgery, which demand deep, specialized causal knowledge rather than general physical rules, remains a critical unexplored gap. To systematically address this challenge, we present \textbf{SurgVeo}, the first expert-curated benchmark for video generation model evaluation in surgery, and the \textbf{Surgical Plausibility Pyramid} (SPP), a novel, four-tiered framework tailored to assess model outputs from basic appearance to complex surgical strategy. On the basis of the SurgVeo benchmark, we task the advanced Veo-3 model with a zero-shot prediction task on surgical clips from laparoscopic and neurosurgical procedures. A panel of four board-certified surgeons evaluates the generated videos according to the SPP. Our results reveal a distinct "plausibility gap": while Veo-3 achieves exceptional Visual Perceptual Plausibility, it fails critically at higher levels of the SPP, including Instrument Operation Plausibility, Environment Feedback Plausibility, and Surgical Intent Plausibility. This work provides the first quantitative evidence of the chasm between visually convincing mimicry and causal understanding in surgical AI. Our findings from SurgVeo and the SPP establish a crucial foundation and roadmap for developing future models capable of navigating the complexities of specialized, real-world healthcare domains.

}

\date{October 30, 2025}
\correspondence{\texttt{zchen.francis@gmail.com}}

\begin{document}
\thispagestyle{firstheader}
\maketitle
\pagestyle{empty}

\section{Introduction}\label{sec:introduction}

The pursuit of Artificial General Intelligence (AGI) hinges on developing systems that can understand and interact with the world in flexible, generalizable ways rather than excelling at narrow, predefined tasks \cite{ha2018world}. A fundamental challenge in this pursuit is enabling machines to build internal representations of how the world works, \textit{i.e.}, understanding not just what is observed, but how environments evolve, how actions lead to consequences, and how future states can be anticipated and reasoned about \cite{lake2017building}. This challenge has catalyzed fundamental research into world models, as learning representations that encode environmental dynamics and enable prediction, planning, and reasoning. Serving as a cornerstone for AGI systems, world models capture the underlying principles governing how the world evolves in response to actions and interventions \cite{lecun2022path,bai2025masks,he2025bridging,zhang2025world}. These models are not merely passive observers but active simulators capable of anticipating future states, evaluating potential outcomes, and supporting decision-making across diverse contexts \cite{hafner2019learning,hafner2019dream}. The ability to build accurate world models represents a critical step toward systems that can understand, interact with, and reason about complex environments in ways that approach human-level intelligence. 

Recent advances in video generation have demonstrated a paradigm shift in the pursuit of world models, with state-of-the-art models demonstrating remarkable capabilities that position them as potential general-purpose world simulators \cite{brooks2024video,wiedemer2025video,he2025pre}. For instance, Veo-3 \cite{wiedemer2025video}, trained on large-scale video data, introduces the Chain-of-Frames (CoF) reasoning mechanism, which parallels chain-of-thought reasoning in language models but operates across the spatiotemporal dimensions of the real world. While language models manipulate human-invented symbols through sequential reasoning steps \cite{sanderson2023gpt,guo2025deepseek}, video generation models apply changes frame-by-frame across time and space, enabling step-by-step visual reasoning that tackles challenging problems requiring temporal and spatial understanding. Through this sophisticated CoF approach integrated with advanced spatiotemporal architectures, these models exhibit emergent physical understanding in natural scenes \cite{wiedemer2025video}. In particular, they perceive 3D spatial relationships and object properties, model temporal dynamics and long-range coherence, manipulate object interactions and material responses, and reason about physical causality and event sequences, thereby achieving photorealistic video generation with temporal consistency, maintaining object permanence, realistic physics including gravity and collision dynamics, and coherent camera motion with proper parallax effects. These impressive results have garnered significant attention across diverse research communities, highlighting the potential of video generation models as foundation world models for understanding and simulating the physical world \cite{reed2022generalist,yuan2025magictime}.

Despite the potential of video generation models as world simulators and their remarkable performance in natural scenes, their application to healthcare contexts remains underexplored, a critical gap given that medical domains require not only visual perception but also deep expert knowledge fundamentally different from general physical understanding \cite{moor2023foundation}. Healthcare is characterized by complex causal relationships involving physiological processes, anatomical structures, and biological mechanisms, where interventions produce outcomes governed by intricate interactions between tissues, organs, and therapeutic agents. Understanding and predicting these dynamics demands specialized domain expertise spanning anatomy, pathology, biomechanics, and clinical reasoning accumulated through years of rigorous training. Surgical procedures represent a particularly demanding testbed for evaluating whether world models can transcend common-sense reasoning to capture this specialized knowledge. Surgery epitomizes the complexity of medical practice, involving real-time dynamic interactions between surgical instruments, anatomical tissues, and biological systems under precise spatiotemporal constraints \cite{varghese2024artificial,chen2025artificial,long2025surgical,zhai2024artificial,yuan2024procedure}. Unlike diagnostic imaging \cite{liu2024most} or treatment planning \cite{chen2025map}, where static or slowly-evolving states dominate, surgical contexts require understanding rapid instrument movements, tissue deformation and biomechanical responses, fluid dynamics of bleeding and irrigation, and the intricate cause-and-effect relationships between instrument actions and anatomical changes governed by expert procedural knowledge \cite{taylor2003medical,chen2023surgical,luo2024surgplan,chen2024surgfc,wang2026endochat,wu2024surgbox}. The gap between the general physical rules learned from natural videos and the specialized knowledge required for high-stakes surgical domains poses a fundamental question: \textit{can current world models bridge this divide to achieve not merely visual plausibility but surgical plausibility necessary for meaningful clinical applications?}

\begin{figure*}[!t]
	\begin{center}
		\includegraphics[width=1\linewidth]{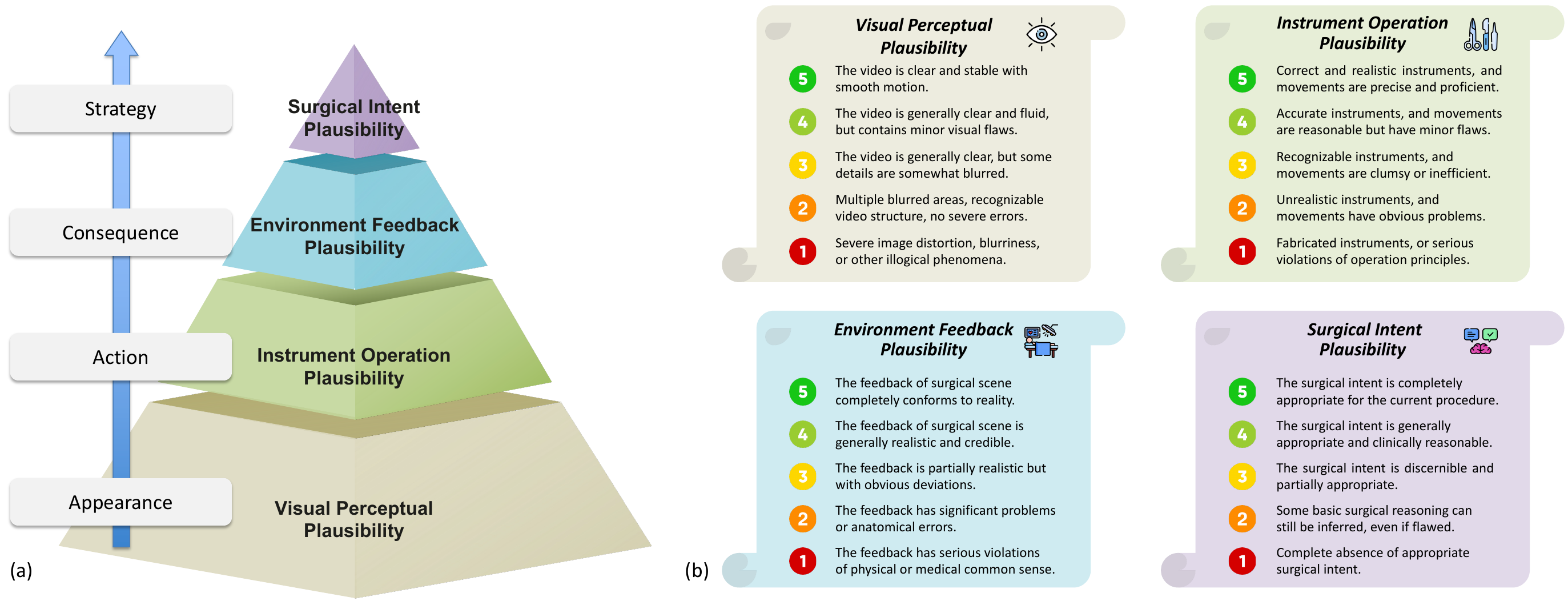}
	\end{center}
	\caption{(a) The Surgical Plausibility Pyramid (SPP) framework, illustrating four hierarchical assessment dimensions: (i) Visual Perceptual Plausibility at the appearance level, assessing the clarity and stability of generated videos, (ii) Instrument Operation Plausibility at the action level, judging the accuracy and technical proficiency of instrument manipulation, (iii) Environment Feedback Plausibility at the consequence level, measuring the realism and credibility of scene feedback, and (iv) Surgical Intent Plausibility at the Strategy level, evaluating the appropriateness and clinical reasoning of surgical actions. (b) Detailed 5-point scoring rubrics (5=excellent to 1=poor) for evaluating each dimension.}
	\label{fig:pyramid}
\end{figure*}

To answer this fundamental question, we present the first systematic evaluation of a state-of-the-art video generation model's potential to serve as a world model in the surgical domain. We introduce the \textbf{SurgVeo} benchmark, a novel dataset curated specifically for this task. It comprises 50 video clips sourced from six independent recordings of laparoscopic hysterectomy \cite{wang2022autolaparo} and endoscopic pituitary surgery \cite{pitvis}, covering a diverse range of procedural stages and complexities to ensure a representative testbed. Using this benchmark, we task the Veo-3 model \cite{wiedemer2025video} with generating 8-second video continuations from a single input frame under two prompting conditions: a baseline prompt and a stage-aware prompt. To move beyond superficial visual metrics and enable a deeper, clinically-grounded assessment, we devise the \textbf{Surgical Plausibility Pyramid} (SPP), a four-tiered evaluation hierarchy designed to dissect a model's capabilities, as elaborated in Fig. \ref{fig:pyramid}. The SPP progresses from the concrete to the abstract: (1) \textbf{Visual Perceptual Plausibility}, assessing the basic appearance of the scene, (2) \textbf{Instrument Operation Plausibility}, evaluating the physical action of instruments, (3) \textbf{Environment Feedback Plausibility}, measuring the causal consequence of actions on tissue, and (4) \textbf{Surgical Intent Plausibility}, examining the overarching strategy behind the procedure. On the basis of SPP, the SurgVeo benchmark is assessed by a panel of four board-certified surgeons, as illustrated in Fig. \ref{fig:pipeline}.

Our results reveal that Veo-3 achieves high scores in visual perceptual plausibility, demonstrating its remarkable capability in generating photorealistic surgical imagery, with some outputs even surprising expert surgeons. However, the assessment exhibits significant shortcomings in higher-level domains, such as Instrument Operation Plausibility and Environment Feedback Plausibility, failing to produce realistic instrument trajectories and biomechanical tissue responses. Additionally, in terms of Surgical Intent Plausibility, Veo-3 struggles to infer the intent behind surgical actions based on the input video and stage information, highlighting its inability to recognize procedural goals. These findings expose a critical gap between generating visually convincing videos and capturing the specialized causal knowledge essential for expert-level understanding in high-stakes surgical contexts. On the basis of the SurgVeo benchmark and the SPP, our professional evaluations provide valuable insights into the challenges in developing surgical world models that bridge general physical reasoning and domain-specific expertise. Our results have significant implications for clinical applications, including surgical training, preoperative planning, intraoperative guidance systems, and autonomous surgical robotics. To foster further research, we will publicly release the SurgVeo benchmark and feedback from expert surgeons.

\section{The SurgVeo Benchmark}
\begin{figure*}[!t]
	\begin{center}
		\includegraphics[width=1\linewidth]{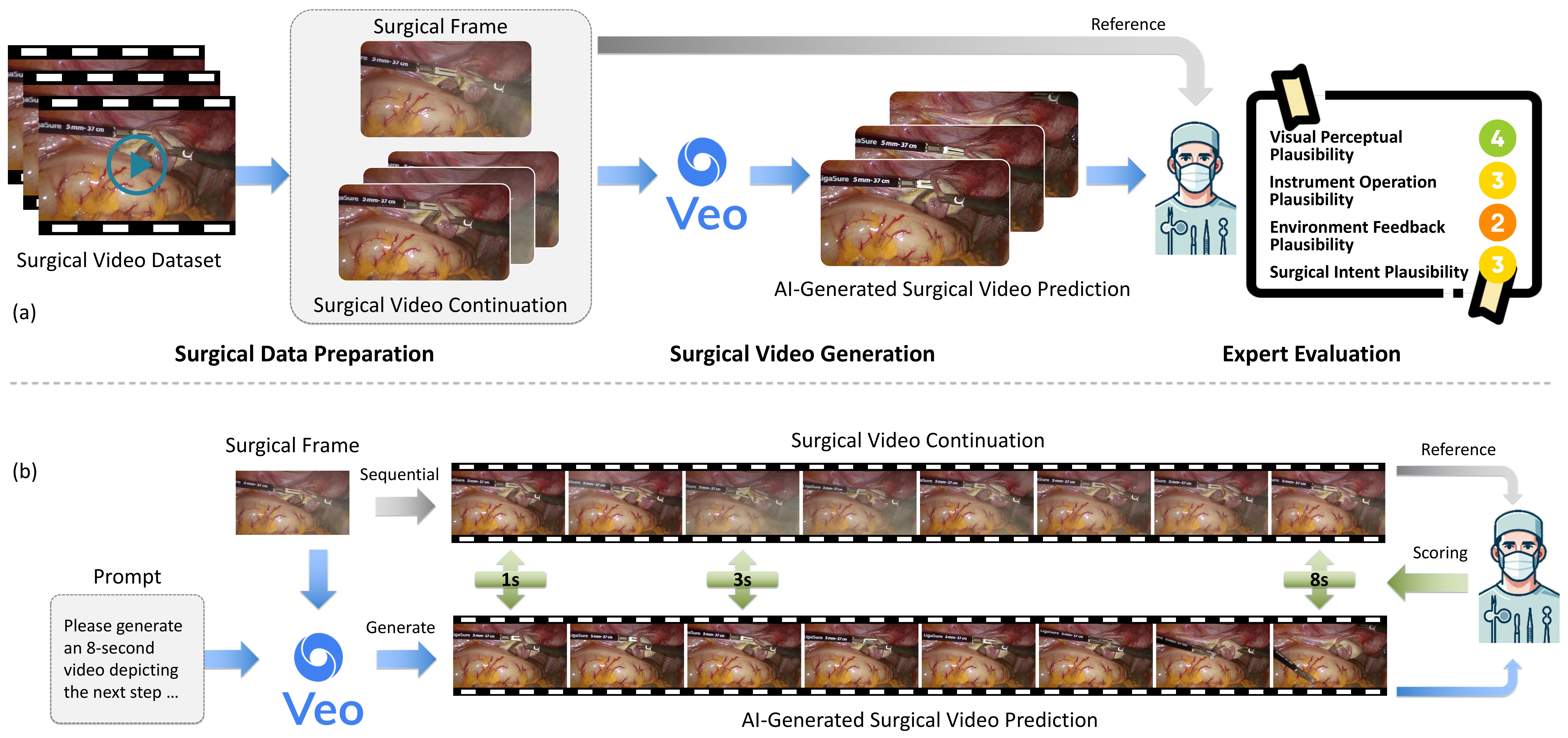}
	\end{center}
	\caption{The overall pipeline of this study. (a) The overview of the SurgVeo benchmark preparation and evaluation workflow. The surgical video dataset is processed to create the paired surgical frame and surgical video continuation. The Veo model takes the surgical frame with a prompt as input to generate the surgical video prediction. A panel of four board-certified surgeons evaluates the generated surgical videos against the real surgical video continuation as reference under the Surgical Plausibility Pyramid (SPP). (b) The illustration of the generation and evaluation process for a single sample in the SurgVeo benchmark. A starting surgical frame and a text prompt are fed into the Veo model to generate an 8-second surgical video prediction. This output is then scored by expert surgeons by comparing it to the real 8-second reference video with a focus on four dimensions of surgical plausibility, particularly at the 1-second, 3-second, and 8-second time points.}
	\label{fig:pipeline}
\end{figure*}

To systematically evaluate the capabilities of state-of-the-art video generation models as \textbf{world models} in the surgical domain, we curate the \textbf{SurgVeo} benchmark and conduct a comprehensive assessment with the help of a panel of four surgeon experts. This study mainly comprises three core components, including (1) the surgical data preparation, (2) a standardized zero-shot video generation task, and (3) a rigorous, multi-dimensional evaluation protocol executed by four board-certified surgeons. The pipeline of this study is summarized in Fig. \ref{fig:pipeline}

\subsection{Surgical Data Preparation}


We introduce the \textbf{SurgVeo} benchmark, the first publicly available benchmark specifically designed for evaluating the world modeling capabilities of video generation models in surgical contexts. Unlike existing video generation benchmarks that focus on natural scenes \cite{huang2024vbench} or general-purpose video quality metrics \cite{hou2024training}, the SurgVeo benchmark is tailored to assess whether state-of-the-art video generation models can capture the specialized dynamics, causal relationships, and domain-specific knowledge inherent in surgical procedures. 

\subsubsection{Rationale and Selection of Surgical Procedures}
The primary objective of the SurgVeo benchmark is to assess the capability of the video generation models to comprehend complex causal relationships within the specialized surgery domain. To this end, we select two representative yet distinct surgical procedures that present unique challenges: laparoscopic hysterectomy \cite{wang2022autolaparo} and endoscopic pituitary surgery \cite{pitvis}. For the laparoscopic surgery track, laparoscopic hysterectomy is a common minimally invasive procedure, and represents the challenges of endoscopic camera views, including instrument interaction in a confined space, soft tissue deformation, and adherence to a specific surgical workflow. For the neurosurgery track, endoscopic pituitary surgery is a high-precision neurosurgical procedure, and represents the demand for simulating delicate, high-stakes maneuvers, including navigating fine anatomical structures from an endoscopic view, avoiding critical neurovascular bundles, and modeling extremely subtle instrument manipulations.

\subsubsection{Data Sourcing and Structure}
To ensure data diversity and representativeness, we source full-length, high-definition surgical video recordings from established public datasets. For the laparoscopic surgery track, we collect 3 independent laparoscopic hysterectomy surgeries from the AutoLaparo dataset \cite{wang2022autolaparo} with the resolution of $1,920\times1,080$ at 25 frames-per-second (fps). For the neurosurgery track, we adopt 3 independent endoscopic pituitary surgeries from the PitVis dataset \cite{pitvis} with the resolution of $1,280\times 720$ at 24 fps.

From these surgical video recordings, we curate surgical video clips that comprehensively cover the procedural workflow. The laparoscopic hysterectomy is categorized into seven distinct stages, including the preparation, dividing the ligament and peritoneum, dividing the uterine vessels and ligament, transecting the vagina, specimen removal, suturing, and washing. The endoscopic pituitary surgery is segmented into twelve stages, including the nasal corridor creation, anterior sphenoidotomy, septum displacement, sphenoid sinus clearance, hemostasis, sellotomy, durotomy, tumor excision, synthetic graft placement, fat graft placement, dural sealant, and debris clearance.

In total, the SurgVeo benchmark comprises 50 distinct surgical clips. As such, we process each surgical clip into a sample in the SurgVeo benchmark, which consists of two components, including the input frame and the referenced surgical video. In particular, the input frame captures the surgical scene immediately preceding a procedural action, and the referenced surgical video, from the original surgical recording that immediately follows the input frame, serves as the reference for expert evaluation. Detailed information about the benchmark tracks, statistics, and data structure can be found in Appendix~\ref{sec:benchmark-track}, Appendix \ref{sec:benchmark-statistics} and Appendix~\ref{sec:data-structure}.

In this way, the inclusion of different surgical specialties, multiple independent procedures, and comprehensive stage coverage ensures the representativeness and diversity of the SurgVeo benchmark. To foster further research and enable rigorous comparison of future models, the complete SurgVeo benchmark will be publicly released (see Appendix~\ref{sec:public-release-reproducibility}). 

\subsection{Surgical Video Generation}


\subsubsection{Zero-Shot Video Generation Task}
The core task of this study is to evaluate the zero-shot surgical video generation of the advanced video generation models. The Veo-3 model is adopted without any fine-tuning on surgical video data, and is tasked to generate a logically coherent and surgically plausible 8-second video continuation based solely on the input frame and a text prompt.

This formulation tests whether video generation models have internalized sufficient understanding of surgical dynamics, instrument behavior, tissue responses, and procedural logic to extrapolate plausible future states from limited visual information. The input frame constraint is deliberately challenging, designed to assess whether models possess genuine surgical knowledge rather than simply extending patterns observable in longer contexts. The 8-second prediction horizon requires maintaining temporal coherence, physical plausibility, and surgical logic over a clinically meaningful duration, encompassing complete surgical actions such as tissue dissection sequences, instrument exchanges, or hemostatic maneuvers.

\subsubsection{Prompting Strategy}
To investigate the influence of procedural knowledge on the video quality of video generation models, we design two distinct prompting conditions, including the baseline prompt and the stage-aware prompt. 

\begin{enumerate}
\item Baseline Prompt: Provide only the procedure type (\textit{e.g.}, "This is a laparoscopic hysterectomy") and general requirements for visual quality and anatomical realism. This condition tests whether the model can autonomously infer appropriate next steps from visual context alone. 

\item Stage-aware Prompt: In addition to baseline information, we further specify the current surgical stage explicitly (\textit{e.g.}, "The current stage is vessel ligation"). This condition tests whether providing explicit contextual knowledge improves the surgical plausibility of the generated video. 
\end{enumerate}

This dual-prompting design enables systematic assessment of whether current video generation models can recognize surgical stages from visual information or require additional textual guidance, a distinction critical for understanding the depth of their learned surgical knowledge. Complete details of the task formulation, prompting strategy, and specific prompt templates used for both surgical tracks are provided in Appendix~\ref{sec:generation-task-formulation}, and Appendix~\ref{sec:prompt-templates}.

\subsection{Expert Evaluation}


A distinguishing strength of this work lies in the rigorous expert evaluation protocol. This expert-driven approach is critical because it transcends superficial visual quality metrics to examine whether generated videos demonstrate a genuine understanding of surgical reasoning, instrument operation, and tissue biomechanics.

\subsubsection{Evaluation Panel}
The evaluation is conducted by a panel of four board-certified surgeons with extensive clinical experience. To ensure domain-specific expertise, two laparoscopic surgeons independently evaluate the laparoscopic hysterectomy videos of the laparoscopic surgery track, and two neurosurgeons independently evaluate the endoscopic pituitary surgery videos of the neurosurgery track. We posit that human expertise is indispensable for accurately assessing the subtle yet critical aspects of surgical plausibility, a nuance that cannot be captured by automated metrics.

\subsubsection{Surgical Plausibility Pyramid}
We propose the Surgical Plausibility Pyramid (SPP) framework to structure the evaluation, with four hierarchical dimensions progressing from concrete visual attributes to abstract strategic reasoning. From the base to the apex of the pyramid, these four hierarchical dimensions are summarized as follows:

\begin{enumerate}
\item  \textbf{Visual Perceptual Plausibility}: Assess the fundamental visual quality and appearance of the generated surgical scene, including clarity, lighting, tissue texture realism, and video smoothness.

\item \textbf{Instrument Operation Plausibility}: Evaluate the physical action, focusing on whether the changes, trajectories, and manipulation techniques of surgical instruments are technically sound and physically plausible.

\item  \textbf{Environment Feedback Plausibility}: Measure the direct consequence of the action, evaluating whether the responses of tissues and organs (\textit{e.g.}, deformation, bleeding) conform to biomechanical and anatomical principles.

\item \textbf{Surgical Intent Plausibility}: Examine the highest level of abstraction, the underlying strategy, by assessing whether the predicted sequence of actions demonstrates a clear, logical, and stage-appropriate objective.
\end{enumerate}

In essence, these dimensions provide a multi-level assessment of the generated video, evaluating its appearance, the plausibility of the depicted action, the realism of its consequence, and the coherence of the underlying surgical strategy, as elaborated in Fig. \ref{fig:pyramid}. We further demonstrate the detailed scoring criteria of the SPP dimensions in Appendix~\ref{sec:expert-evaluation-protocol}. 

For the evaluation, surgeons use the 8-second real surgical video as a professional reference to understand the context and what a correct surgical progression entails. With this reference in mind, they then independently score the 8-second generated video, providing scores on a 5-point integer scale (1 = Very Poor; 5 = Indistinguishable from Reality) at three specific temporal checkpoints: 1-second, 3-second, and 8-second. This temporal evaluation structure allows for tracking the degradation or maintenance of plausibility over the prediction horizon. Taken together, this multi-faceted evaluation protocol establishes a robust and reproducible framework for quantifying the gap between visual realism and true surgical understanding in advanced video generation models, providing a critical tool for future research in this domain.

\section{Results}





\begin{table}[t]
\centering
\caption{Evaluation scores of SurgVeo benchmark on the laparoscopic surgery track across different time points and prompt strategies. Scores are reported on a 1-5 scale, where 5 is the best, and the standard deviation is calculated with the average score of two laparoscopic surgery experts.}
\label{tab:evaluation_scores_laparo}
\begin{tabular}{c|c|cccc}
\toprule
\makecell[c]{\textbf{Prompt}\\\textbf{Strategy}} & \makecell[c]{\textbf{Time}\\\textbf{Point}} & \makecell[c]{\textbf{Visual Perceptual}\\\textbf{Plausibility}} & \makecell[c]{\textbf{Instrument Operation}\\\textbf{Plausibility}} & \makecell[c]{\textbf{Environment Feedback}\\\textbf{Plausibility}} & \makecell[c]{\textbf{Surgical Intent}\\\textbf{Plausibility}} \\ 
\midrule
\multirow{3}{*}{\makecell[c]{Baseline\\Prompt}} & 1 & 3.72 ± 0.24 & 3.36 ± 0.20 & 3.06 ± 0.08 & 3.11 ± 0.16 \\
                    & 3  & 3.69 ± 0.20 & 2.33 ± 0.16 & 2.06 ± 0.24 & 2.03 ± 0.35 \\
                    & 8  & 3.56 ± 0.31 & 1.78 ± 0.00 & 1.64 ± 0.12 & 1.61 ± 0.16 \\
\hline
\multirow{3}{*}{\makecell[c]{Stage-aware\\Prompt}} & 1  & 3.61 ± 0.24 & 3.22 ± 0.16 & 3.11 ± 0.24 & 3.22 ± 0.00 \\
                    & 3  & 3.58 ± 0.27 & 2.31 ± 0.04 & 2.08 ± 0.04 & 2.11 ± 0.24 \\
                    & 8  & 3.39 ± 0.47 & 1.69 ± 0.20 & 1.53 ± 0.12 & 1.81 ± 0.20 \\
\bottomrule
\end{tabular}
\end{table}

Our experimental results reveal a profound disconnect between Veo-3's capacity for visual synthesis and its understanding of surgical knowledge. The detailed evaluation scores and score distributions, presented for the laparoscopic surgery track in Table \ref{tab:evaluation_scores_laparo} and Fig. \ref{fig:violin_laparo} and for the neurosurgery track in Table \ref{tab:evaluation_scores_neuro} and Fig. \ref{fig:violin_neuro}, provide comprehensive quantitative evidence for this gap. In addition, we present a case-by-case analysis of representative high-scoring and low-scoring examples from the SurgVeo benchmark in Appendix \ref{sec:good-bad-examples}, contextualized with synthesized expert feedback. The findings are organized below by an analysis of performance across the Surgical Plausibility Pyramid, the temporal degradation of plausibility, discrepancies between surgical specialties, the ineffectiveness of prompting strategies, and a detailed qualitative and quantitative video error analysis.

\subsection{Performance Across the Surgical Plausibility Pyramid}

Our primary finding is a stark dichotomy in the performance of the generated surgical videos, clearly illustrated across both surgical types (Table \ref{tab:evaluation_scores_laparo} and Table \ref{tab:evaluation_scores_neuro}). The Veo-3 consistently excels at the base of the Surgical Plausibility Pyramid, achieving high scores in Visual Perceptual Plausibility. For both laparoscopic and neurosurgical procedures, the mean initial scores for this dimension are high (\textit{e.g.}, the baseline prompt achieves 3.72 ± 0.24 and 3.88 ± 0.09, respectively). In addition, the violin plots (Fig. \ref{fig:violin_laparo} and Fig. \ref{fig:violin_neuro}) visually confirm this: the score distributions for this dimension are tightly clustered at the high end of the scale (\textit{i.e.}, mostly between 3.0 and 5.0). Surgeons note the imagery is often "shockingly clear" and texturally realistic.

However, this quality collapses when assessed against the higher levels of the pyramid, which require causal understanding. The mean scores in Table \ref{tab:evaluation_scores_laparo} and Table \ref{tab:evaluation_scores_neuro} for Instrument Operation, Environment Feedback, and Surgical Intent Plausibility are much lower, often falling below 2.0. In laparoscopic procedures with stage-aware prompt (Table \ref{tab:evaluation_scores_laparo}), the average scores for Instrument Operation (1.69 ± 0.20), Environment Feedback (1.53 ± 0.12), and Surgical Intent (1.81 ± 0.20) at the 8-second are all critically low. The deficit is even more pronounced in neurosurgery (Table \ref{tab:evaluation_scores_neuro}), where scores for the same dimensions are consistently poor from the outset (\textit{e.g.}, 1.13 ± 0.04 and 1.17 ± 0.02 at the 8-second with the baseline prompt and stage-aware prompt, respectively). On the other hand, the violin plots in Fig. \ref{fig:violin_laparo} and Fig. \ref{fig:violin_neuro} dramatically visualize this failure: the distributions for these three dimensions are heavily skewed, with the vast majority of scores concentrated at the bottom of the scale (between 1.0 and 2.5). This "plausibility gap", the difference between high visual quality and low surgical logic, is the central quantitative finding of our study.

\begin{table}[t]
\centering
\caption{Evaluation scores of SurgVeo benchmark on neurosurgery track across different time points and prompt strategies. Scores are reported on a 1-5 scale, where 5 is the best and the standard deviation is calculated with the average score of two neurosurgery experts.}
\label{tab:evaluation_scores_neuro}
\begin{tabular}{c|c|cccc}
\toprule
\makecell[c]{\textbf{Prompt}\\\textbf{Strategy}} & \makecell[c]{\textbf{Time}\\\textbf{Point}} & \makecell[c]{\textbf{Visual Perceptual}\\\textbf{Plausibility}} & \makecell[c]{\textbf{Instrument Operation}\\\textbf{Plausibility}} & \makecell[c]{\textbf{Environment Feedback}\\\textbf{Plausibility}} & \makecell[c]{\textbf{Surgical Intent}\\\textbf{Plausibility}} \\ 
\midrule
\multirow{3}{*}{\makecell[c]{Baseline\\Prompt}} & 1 & 3.88 ± 0.09 & 2.77 ± 0.02 & 2.84 ± 0.09 & 2.03 ± 0.09 \\
                    & 3  & 3.53 ± 0.22 & 2.08 ± 0.11 & 2.16 ± 0.00 & 1.42 ± 0.20 \\
                    & 8  & 3.41 ± 0.22 & 1.75 ± 0.04 & 1.78 ± 0.18 & 1.13 ± 0.04 \\
\hline
\multirow{3}{*}{\makecell[c]{Stage-aware\\Prompt}} & 1  & 3.84 ± 0.04 & 2.58 ± 0.07 & 2.64 ± 0.02 & 1.97 ± 0.13 \\
                    & 3  & 3.39 ± 0.07 & 2.02 ± 0.07 & 2.00 ± 0.00 & 1.42 ± 0.07 \\
                    & 8  & 3.25 ± 0.09 & 1.75 ± 0.04 & 1.73 ± 0.07 & 1.17 ± 0.02 \\
\bottomrule
\end{tabular}
\end{table}

\subsection{Temporal Degradation of Plausibility}
Analysis of the scoring data across time points reveals a consistent and significant degradation in surgical plausibility as the prediction horizon extends. This trend is detailed in the mean scores of Table \ref{tab:evaluation_scores_laparo} and Table \ref{tab:evaluation_scores_neuro} and visualized in the score distributions of Fig. \ref{fig:violin_laparo} and Fig. \ref{fig:violin_neuro}. A critical observation from the violin plots (Fig. \ref{fig:violin_laparo} and Fig. \ref{fig:violin_neuro}) is the differential stability of the SPP dimensions over time. For both surgery types and both prompting strategies, the score distributions for Visual Perceptual Plausibility remain relatively high and stable across the 1, 3, and 8-second time points. In contrast, the distributions for the three higher-level dimensions, Instrument Operation, Environment Feedback, and Surgical Intent Plausibility, show a clear and significant downward trend, with the violins becoming progressively more compressed at the bottom of the scale as time progresses.

This is numerically evident in the mean scores. For instance, in the laparoscopic evaluations (Table \ref{tab:evaluation_scores_laparo} with the baseline prompt), the score for Environment Feedback Plausibility plummeted from 3.06 ± 0.08 at 1 second to 1.64 ± 0.12 by 8 seconds, a drop of nearly 46\%. Similarly, Surgical Intent Plausibility fell from 3.11 ± 0.16 to 1.61 ± 0.16. This pattern is consistent with our central SPP conclusion: the Veo model excels at generating basic visual appearance, which it can maintain, but fails at complex surgical logic. As the prediction horizon increases, the difficulty of maintaining this logic compounds, leading to an accumulation of predictive errors and a catastrophic failure in long-range coherence.

\subsection{Discrepancies Between Surgical Specialties}
\begin{figure*}[t]
	\begin{center}
		\includegraphics[width=1\linewidth]{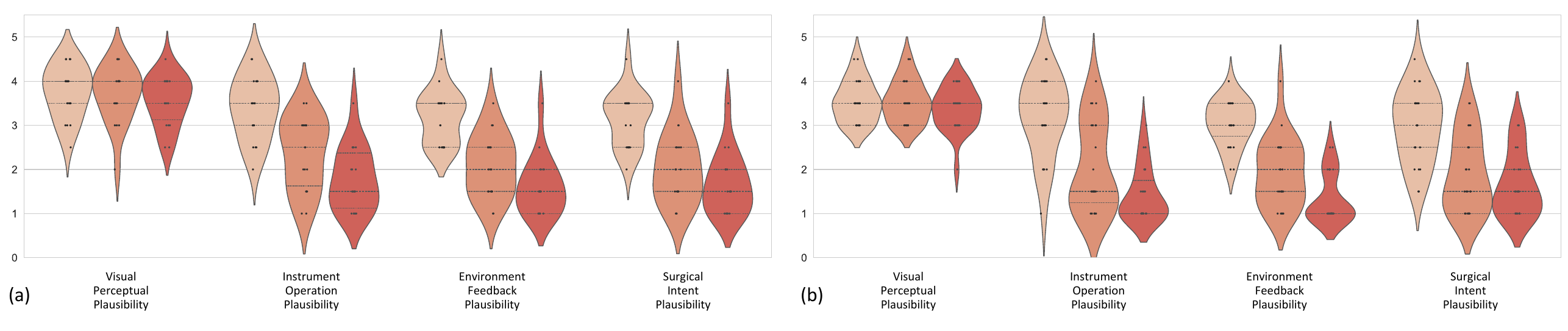}
	\end{center}
	\caption{Violin plots illustrating the performance on the laparoscopic surgery track in the SurgVeo benchmark. Results are shown for (a) the baseline prompt and (b) the stage-aware prompt. The performance is assessed across four evaluation dimensions in the SPP, with three progressively deeper shades representing evaluations at 1-second, 3-second, and 8-second. Each sample point reflects the average score provided by two laparoscopic surgery experts.}
	\label{fig:violin_laparo}
\end{figure*}

\begin{figure*}[t]
	\begin{center}
		\includegraphics[width=1\linewidth]{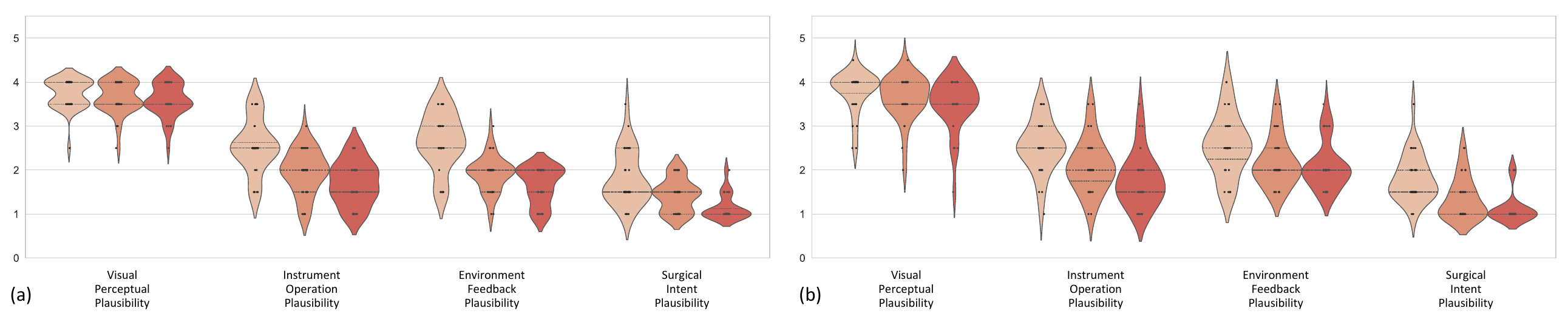}
	\end{center}
    \caption{Violin plots illustrating the performance on the neurosurgery track in the SurgVeo benchmark. Results are shown for (a) the baseline prompt and (b) the stage-aware prompt. The performance is assessed across four evaluation dimensions in the SPP, with three progressively deeper shades representing evaluations at 1-second, 3-second, and 8-second. Each sample point reflects the average score provided by two neurosurgery experts.}
	\label{fig:violin_neuro}
\end{figure*}

While the performance of high-level plausibilities is poor across both surgical specialties, the results indicate that simulating neurosurgery is a more challenging task. This is evident not only in the mean scores from Table \ref{tab:evaluation_scores_laparo} and Table \ref{tab:evaluation_scores_neuro} but also in the score distributions shown in Fig. \ref{fig:violin_laparo} and Fig. \ref{fig:violin_neuro}. A direct comparison shows that the plausibility scores for neurosurgery are typically lower since the very first second of generation. For example, under the baseline prompt at the 1-second, Instrument Operation Plausibility is rated 3.36 ± 0.20 for laparoscopy but only 2.77 ± 0.02 for neurosurgery. The violin plots reinforce this: the distributions for the three higher-level plausibility dimensions in the neurosurgery track (Fig. \ref{fig:violin_neuro}) are even more severely compressed at the bottom of the scale than their counterparts in the laparoscopic surgery track (Fig. \ref{fig:violin_laparo}). This supports our hypothesis that the heightened precision, delicate tissue handling, and microscopic scale of neurosurgery present a more complex set of implicit "rules" that the Veo model fails to capture.

\subsection{Ineffectiveness of Stage-Aware Prompting}
Our comparative analysis between prompting strategies reveals that providing additional contextual information yields no significant or consistent improvement in surgical plausibility. Across both Table \ref{tab:evaluation_scores_laparo} and Table \ref{tab:evaluation_scores_neuro}, the scores for the stage-aware prompt are not meaningfully higher than those for the baseline prompt. In several instances, such as for Instrument Operation Plausibility in laparoscopy at the 1-second, the stage-aware score (3.22 ± 0.16) is even lower than the baseline (3.36 ± 0.20). The violin plots in Fig. \ref{fig:violin_laparo} (a vs. b) and Fig. \ref{fig:violin_neuro} (a vs. b) make this finding visually irrefutable. A side-by-side comparison of the baseline and stage-aware plots for each surgery type shows similar distributions for all plausibility dimensions. This finding strongly reinforces our conclusion that the Veo model's limitations arise from a fundamental inability to reason about domain-specific knowledge, rather than a mere lack of context.

\begin{figure*}[!t]
	\begin{center}
		\includegraphics[width=0.99\linewidth]{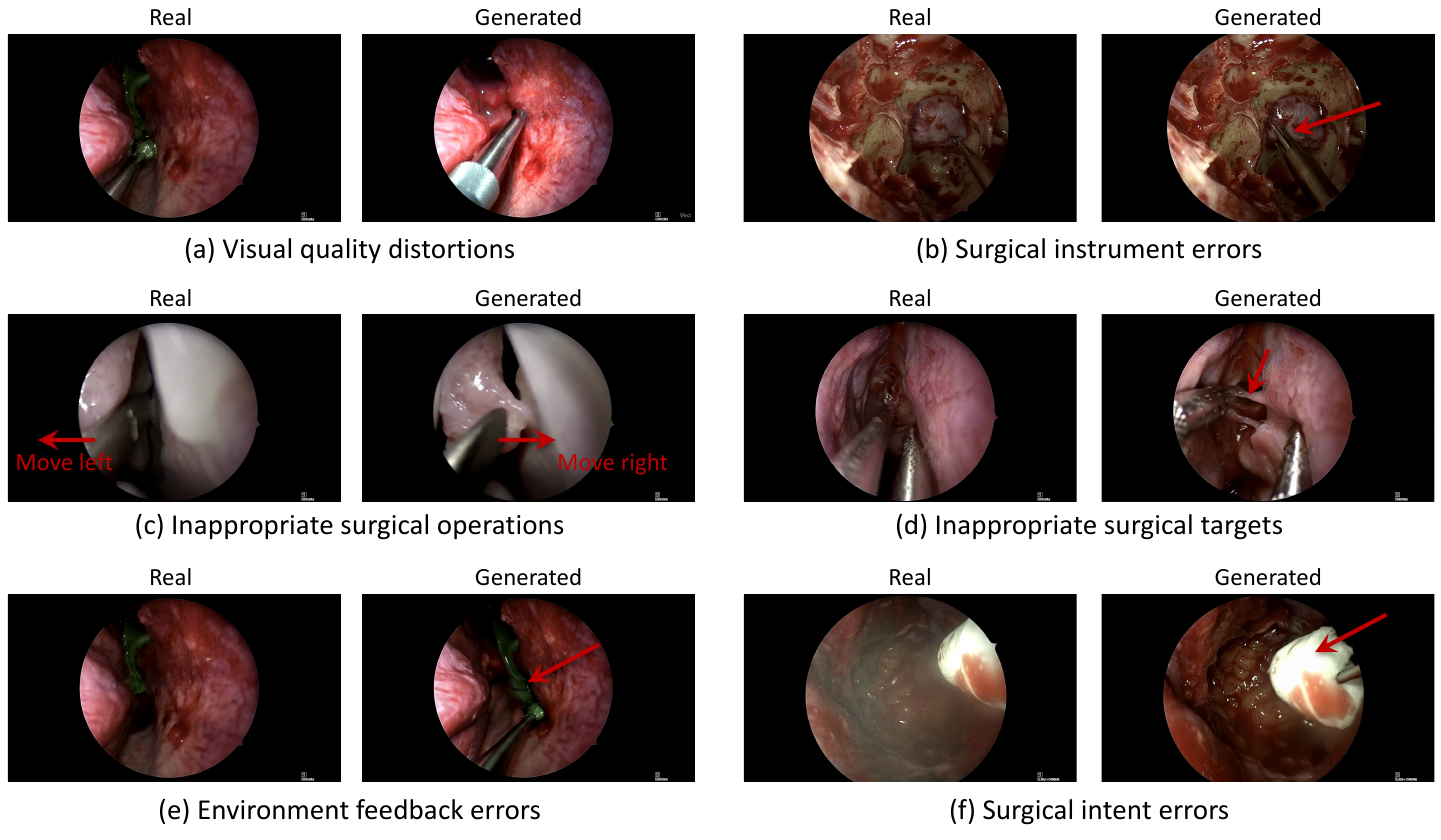}
	\end{center}
	\caption{Qualitative examples of typical failures identified in the generated videos. Each example presents a side-by-side comparison of the real surgical frame (left) and the generated surgical frame (right). These examples elaborate on failures across the Surgical Plausibility Pyramid, including: (a) visual quality distortions, (b) surgical instrument errors, (c) inappropriate surgical operations, (d) inappropriate surgical targets, (e) environment feedback errors, and (f) surgical intent errors. Red arrows indicate specific illogical, anatomically incorrect, or physically impossible artifacts.}
	\label{fig:error_example}
\end{figure*}

\subsection{Video Error Analysis: Qualitative Examples and Quantitative Distribution}\label{section_video_error_analysis}
To provide a comprehensive overview of the Veo model's failures, we compile the primary error categories identified by our four expert surgeons during their evaluation of the entire SurgVeo benchmark. Fig. \ref{fig:error_example} presents typical visual examples of these critical failures, comparing the generated frame to the real-world reference video. These examples visually corroborate the surgeons' qualitative feedback, showcasing specific, critical failures:

\begin{itemize}
    \item Visual quality distortions: The generated video exhibits a sudden, unnatural increase in brightness, inconsistent with stable surgical lighting conditions.

    \item Surgical instrument errors: The model hallucinates a non-existent, fabricated instrument, whereas the real procedure uses a standard scalpel.

    \item  Inappropriate surgical operations: The generated video shows the dissector moving to the right, which is procedurally incorrect for the scene; the real operation requires a leftward movement.

    \item Inappropriate surgical targets: The model depicts an instrument manipulating mucus, while the correct procedure involves a coordinated action of irrigation and suction on a different target.

    \item Environment feedback errors: The model violates physical laws by showing a suction tool pulling an entire block of gel-foam as if it is a solid, attached mass, rather than suctioning loose fluid and debris from its surface.

    \item Surgical intent errors: The real procedure involves injecting biologic glue onto the dura, but the model incorrectly predicts a completely different intent, \textit{i.e.}, wiping the dura with a cotton patty.
\end{itemize}

\begin{figure*}[!t]
	\begin{center}
		\includegraphics[width=0.7\linewidth]{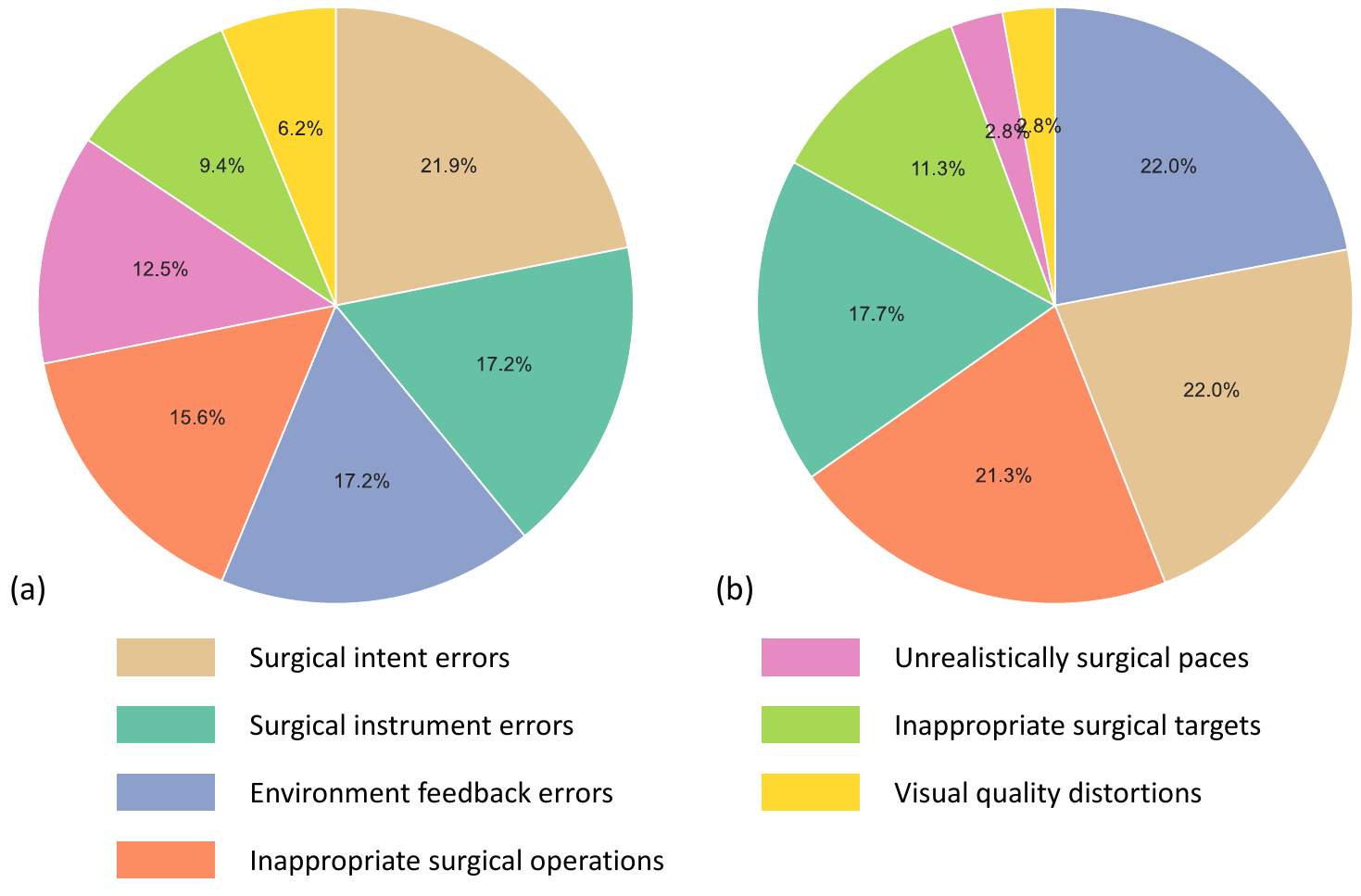}
	\end{center}
	\caption{Distribution of error types identified by expert surgeons in generated videos of the SurgVeo benchmark. The charts quantify the frequency of different failures in surgical plausibility for (a) the laparoscopic surgery track and (b) the neurosurgery track. Across both specialties, errors related to high-level surgical logic, such as Surgical intent errors, Surgical instrument errors, and Inappropriate surgical operations, constitute the vast majority of all failures. In contrast, basic Visual quality distortions represent only a small fraction of the total errors, reinforcing the finding of the plausibility gap.
    }
	\label{fig:pie_error}
\end{figure*}

Then, we quantify the frequency of these error types, with the distribution compiled from the baseline prompt evaluations shown in Fig. \ref{fig:pie_error}. This quantitative analysis provides a striking confirmation of the "plausibility gap". Across both surgery types, basic visual quality distortions account for only a minuscule fraction of all identified failures: 6.2\% in laparoscopic surgery (Fig. \ref{fig:pie_error}(a)) and a mere 2.8\% in neurosurgery (Fig. \ref{fig:pie_error}(b)). Conversely, the vast majority of errors (over 93\% in both cases) are critical failures in surgical logic. In both procedures, the four most dominant error categories are surgical intent errors (21.9\% in laparoscopic surgery, 22.0\% in neurosurgery), environment feedback errors (17.2\% in laparoscopic surgery, 22.0\% in neurosurgery), surgical instrument errors (17.2\% in laparoscopic surgery, 17.7\% in neurosurgery), and inappropriate surgical operations (15.6\% in laparoscopic surgery, 21.3\% in neurosurgery). In neurosurgery (Fig. \ref{fig:pie_error}(b)), these four high-level logical failures alone constitute 83\% of all errors. This quantitative breakdown irrefutably demonstrates that the Veo model's failure is not in visual rendering, but in its fundamental lack of understanding of surgical intent, action, and consequence.

\section{Discussions}


Our study provides a sobering but essential assessment of the current capabilities of video generation models as world models for the specialized domain of surgery. The central finding is the profound "plausibility gap" between the Veo model's ability to render visually convincing surgical scenes and its complete failure to comprehend the underlying causal principles governing them. While Veo-3 can replicate the appearance of surgery with stunning fidelity, it lacks any semblance of surgical understanding, operating as a sophisticated pattern-matcher rather than a knowledgeable simulator.

The ineffectiveness of our stage-aware prompting strategy is a key diagnostic result. Providing the model with explicit contextual information about the surgical stage did not significantly improve plausibility, revealing that the core deficit is not a lack of information but a fundamental inability to process it. The model appears to lack the foundational knowledge required to translate an abstract concept like "vessel ligation" into a concrete, physically sound sequence of actions and consequences. This suggests that the model's internal representations, learned from general-domain videos, are not structured to accommodate the complex, rule-based logic of surgery.

This highlights a critical challenge for the future of world models. While models trained on massive, diverse datasets can learn the "common-sense physics" of the everyday world, surgery operates on a different set of principles: the "expert-sense" of anatomy, physiology, and biomechanics. Our findings suggest that merely scaling up training on general data will be insufficient to bridge this divide. Achieving true world modeling in expert domains will likely require new architectural paradigms capable of integrating structured, domain-specific knowledge and enforcing hard physical and logical constraints on the generative process.

Despite these limitations, the clinical appetite for a true surgical world model is immense. Such a model could revolutionize medical education by providing trainees with interactive, high-fidelity simulators for procedural practice. It could also enhance patient safety by enabling pre-operative rehearsal of complex cases or powering intra-operative guidance systems that monitor for risks and deviations from the optimal surgical plan. Our work, by establishing the SurgVeo benchmark and quantifying the current performance gap, takes a crucial first step toward this future.

The path from the current state-of-the-art to a clinically useful surgical simulator requires targeted research to address the specific deficits identified by SurgVeo. Future efforts should focus on two key directions. First, explicitly incorporating surgical knowledge into world models is paramount. This can leverage existing research in surgical and endoscopic video generation \cite{li2024endora,chen2025surgsora}, providing the necessary foundation for models to learn the rules of surgical practice, instrument usage, and anatomical interactions. Second, advancements in the world model technology itself are needed to better capture physical laws and long-range causal dependencies. Explorations into physics-informed modeling and architectures designed to mitigate catastrophic error accumulation over time will be crucial \cite{yuan2025magictime,PhysWorld}. These dual approaches, enhancing domain-specific knowledge and improving core modeling capabilities, are essential to ensure that generated simulations conform to the complex realities of the operating room and can reliably predict outcomes over extended durations. By providing a rigorous, clinically-grounded framework for evaluation, we hope our study will guide the development of the next generation of generative models, which can transition from simple visual mimicry to genuine causal understanding, unlocking their transformative potential for medicine.

\section{Conclusion}

In this study, we sought to determine how far current state-of-the-art video generation models are from functioning as true world models for surgery. Our findings provide a clear yet sobering answer: a significant gap remains. While these models can master the appearance of surgery with remarkable photorealism, they fundamentally lack an understanding of its practice, failing to adhere to the basic principles of surgical action, consequence, and strategy.

The critical contribution of this work is the establishment of the SurgVeo benchmark and the Surgical Plausibility Pyramid, which together provide a standardized and clinically-grounded framework to measure and guide progress in this challenging domain. This is not a dismissal of the technology's potential but a foundational step and a call to action. By systematically identifying the current limitations, our research provides a clear roadmap for the field to move beyond superficial visual mimicry. The ultimate goal is to develop models with the deep causal understanding necessary to create truly intelligent simulations, thereby unlocking the immense potential of AI to enhance surgical training, planning, and patient safety.

{
\small
\bibliographystyle{IEEEtran}
\bibliography{paper}
}




\newpage
\beginappendix
\appendix

\section{Tracks of SurgVeo Benchmark}
\label{sec:benchmark-track}

The SurgVeo benchmark comprises two distinct tracks to encompass diverse surgical environments and procedural complexities:

\begin{enumerate}
    \item \textbf{Laparoscopic Surgery Track:} This track features video clips from minimally invasive abdominal procedures, characterized by endoscopic camera views, pneumoperitoneum environments, and characteristic instrument configurations including graspers, scissors, and electrocautery devices. Laparoscopic procedures present unique challenges, including a limited field of view, indirect tissue manipulation through long instruments, and specific lighting and depth perception constraints.
    
    \item \textbf{Neurosurgery Track:} This track features video clips from microscopic cranial procedures, characterized by high magnification microscopic views, extremely delicate tissue handling of neural and vascular structures, and specialized neurosurgical instrumentation. Neurosurgical procedures demand exceptional precision, with operations involving critical anatomical structures where millimeter-scale errors can have significant clinical consequences.
\end{enumerate}

\section{Statistics of SurgVeo Benchmark}
\label{sec:benchmark-statistics}

\begin{figure*}[b]
	\begin{center}
		\includegraphics[width=0.9\linewidth]{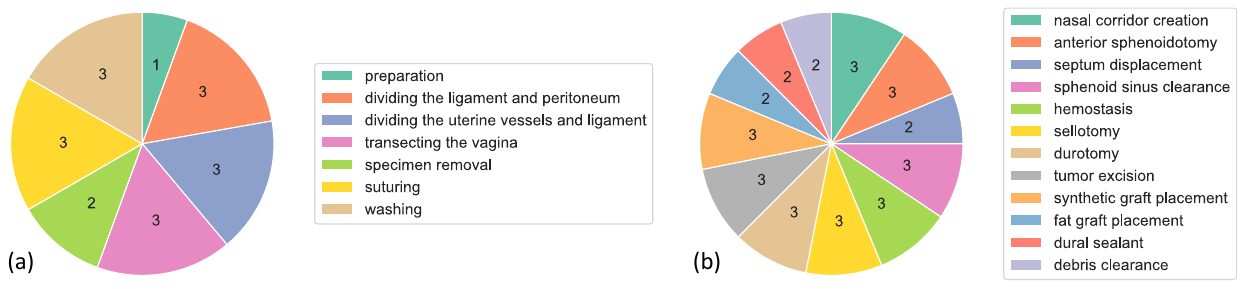}
	\end{center}
	\caption{The surgical stage distribution of the proposed SurgVeo benchmark in two tracks, including (a) the distribution of the laparoscopic surgery track, and (b) the distribution of the neurosurgery track. Note that not every surgery includes all possible surgical stages.}
	\label{fig:surg_distribution}
\end{figure*}

The SurgVeo benchmark dataset consists of a total of 50 surgical video segments, divided into two distinct tracks: the laparoscopic surgery track and the neurosurgery track. This division ensures a diverse representation of surgical procedures, with each track capturing the unique characteristics and key stages of its respective domain. The laparoscopic surgery track focuses on 7 major stages, such as preparation, ligament division, and suturing, while the neurosurgery track covers 12 stages, including nasal corridor creation, tumor excision, and graft placement. Specifically, the laparoscopic surgery track in Fig. \ref{fig:surg_distribution}(a) contains 18 video segments, while the neurosurgery track in Fig. \ref{fig:surg_distribution}(a) includes 32 video segments. Please note that not all surgeries include every stage, reflecting the variability in surgical workflows. In this way, the balanced distribution of videos and stages across both tracks provides a comprehensive dataset for benchmarking surgical scene understanding, enabling robust evaluations of models across different procedures and stages.

\section{Data Structure of SurgVeo Benchmark}
\label{sec:data-structure}

Each sample in the SurgVeo benchmark consists of a continuous surgical video sequence designed to test the model's predictive capabilities:

\begin{itemize}
    \item \textbf{Input Frame:} The first frame of the real video clip capturing the surgical scene immediately preceding a procedural action. This minimal temporal context challenges the model to extrapolate future dynamics from limited information, testing whether it has internalized sufficient surgical knowledge to predict plausible continuations.
    
    \item \textbf{Generated Video (8 seconds):} The model is tasked with generating an 8-second video continuation that predicts the subsequent surgical developments. This extended prediction horizon requires maintaining temporal coherence, physical plausibility, and surgical logic over a clinically meaningful duration.
    
    \item \textbf{Reference Video (8 seconds):} The actual surgical footage following the input frame serves as the reference for expert evaluation. Surgeons compare the AI-generated predictions against these real-world continuations across multiple assessment dimensions.
\end{itemize}

The design philosophy emphasizes \textit{prediction under uncertainty}: given minimal visual context, can the video generation model demonstrate understanding of instrument dynamics, tissue behavior, and surgical intent sufficient to generate plausible future scenarios? This formulation mirrors the cognitive demands placed on surgical trainees learning to anticipate procedural progression.


\section{Task Formulation of Surgical Video Generation}
\label{sec:generation-task-formulation}

The core task is surgical video prediction: given an input frame showing the surgical scene immediately before a procedural action, the Veo-3 model generates an 8-second continuation predicting the subsequent surgical developments. This formulation tests whether the model has internalized sufficient understanding of surgical dynamics, instrument behavior, and tissue responses to extrapolate plausible future states from minimal visual context.

To investigate the impact of procedural context on generation quality, we evaluate Veo-3 under two prompting conditions for each surgical track:

\begin{enumerate}
    \item \textbf{Baseline Prompt for Veo-3:} Provide only the procedure type and general expectations for visual quality and anatomical realism, without specifying the current surgical stage. This condition tests whether the model can infer the appropriate next steps purely from visual context.
    
    \item \textbf{Stage-Aware Prompt for Veo-3:} Explicitly specify the current surgical stage (\textit{e.g.}, ``suturing'' for laparoscopic procedures, ``debris clearance'' for neurosurgical procedures) in addition to the baseline information. This condition tests whether providing explicit procedural context improves the model's ability to generate stage-appropriate actions.
\end{enumerate}

This dual-prompting design allows us to assess whether current video generation models can autonomously recognize surgical stages from visual information alone, or whether they require explicit textual guidance to produce contextually appropriate predictions—a distinction critical for understanding the depth of their learned surgical knowledge.

\section{Prompting Strategy with Templates}
\label{sec:prompt-templates}

\subsection{Laparoscopic Surgery Track}

For the laparoscopic hysterectomy samples in the laparoscopic surgery track, the following prompts are used:

\textbf{Baseline Prompt:}
\begin{quote}
\textit{Please generate an 8-second video depicting the next step in a laparoscopic hysterectomy procedure. The scene should present a realistic surgical environment viewed through the laparoscopic camera, showing clear anatomical structures. The lighting should mimic the cool-light illumination typical of laparoscopic surgery. Ensure that the generated video includes smooth camera adjustments to maintain focus on the surgical site, as well as subtle movements of surrounding tissues caused by instrument interaction. The tissue texture, color, and responsiveness should be anatomically accurate. The overall flow of the procedure must be logical and consistent with real-world surgical practices.}
\end{quote}

\textbf{Stage-Aware Prompt:}
\begin{quote}
\textit{Please generate an 8-second video depicting the next step in a laparoscopic hysterectomy procedure. \textbf{The current stage of the surgery is [STAGE]}. The scene should present a realistic surgical environment viewed through the laparoscopic camera, showing clear anatomical structures. The lighting should mimic the cool-light illumination typical of laparoscopic surgery. Ensure that the generated video includes smooth camera adjustments to maintain focus on the surgical site, as well as subtle movements of surrounding tissues caused by instrument interaction. The tissue texture, color, and responsiveness should be anatomically accurate. The overall flow of the procedure must be logical and consistent with real-world surgical practices.}
\end{quote}
 
\subsection{Neurosurgery Track}

For the endoscopic pituitary surgery samples in the neurosurgery track, the following prompts are used:

\textbf{Baseline Prompt:}
\begin{quote}
\textit{Please generate an 8-second video depicting the next step in an endoscopic pituitary surgery procedure. The scene should present a realistic surgical environment viewed through the endoscopic camera, showing clear anatomical structures. The lighting should mimic the cool-light illumination typical of endoscopic surgery. Ensure that the generated video includes smooth camera adjustments to maintain focus on the surgical site, as well as subtle movements of surrounding tissues caused by instrument interaction. The texture, color, and responsiveness of the tissue should be anatomically accurate. The overall flow of the procedure must be logical and consistent with real-world surgical practices.}
\end{quote}

\textbf{Stage-Aware Prompt:}
\begin{quote}
\textit{Please generate an 8-second video depicting the next step in an endoscopic pituitary surgery procedure. \textbf{The current stage of the surgery is [STAGE]}. The scene should present a realistic surgical environment viewed through the endoscopic camera, showing clear anatomical structures. The lighting should mimic the cool-light illumination typical of endoscopic surgery. Ensure that the generated video includes smooth camera adjustments to maintain focus on the surgical site, as well as subtle movements of surrounding tissues caused by instrument interaction. The texture, color, and responsiveness of the tissue should be anatomically accurate. The overall flow of the procedure must be logical and consistent with real-world surgical practices.}
\end{quote}

\section{Implementation Details of Surgical Video Generation}
\label{sec:technical-implementation}

All video generations are performed using the Veo-3 model accessed through the Google Flow platform\footnote{https://labs.google/fx/tools/flow} with the Veo-3.0-quality version. For each sample in the SurgVeo benchmark, the input frame is provided as a visual prompt, accompanied by the appropriate text prompt (either baseline or stage-aware one, depending on the experimental condition). The Veo-3 model generates 8-second continuation videos at its default resolution and frame rate settings. No additional fine-tuning or domain-specific adaptation is applied to the model, ensuring that our evaluation reflects the zero-shot surgical prediction capabilities of the state-of-the-art video generation model.

\section{Surgical Plausibility Pyramid for Expert Evaluation}
\label{sec:expert-evaluation-protocol}

\subsection{Visual Perceptual Plausibility}
\label{subsec:visual-fidelity}

\textbf{Evaluation Focus:} Comprehensive assessment of the overall visual quality of the video, including image clarity, lighting, tissue texture and color realism, as well as video playback smoothness, motion fluidity, and the presence of artifacts, jitter, or teleportation issues.

\textbf{Scoring Criteria:}
\begin{itemize}
    \item \textbf{5-point:} The video is clear and stable with smooth motion, visually indistinguishable from real high-quality surgical recordings.
    \item \textbf{4-point:} The video is generally clear and fluid, but contains minor visual imperfections, such as occasional subtle jitter in localized areas, slightly blurred texture details, or unnatural lighting transitions. These issues are only noticeable upon careful observation and do not affect the overall viewing experience.
    \item \textbf{3-point:} The video is generally clear, but some details (\textit{e.g.}, tissue reflections) are somewhat blurred, or instruments exhibit slight stuttering during movement.
    \item \textbf{2-point:} The video exhibits obvious quality issues, such as multiple blurred areas, frequent noticeable stuttering or discontinuity in instrument movement, significant color distortion, or visible artifacts. However, the overall video structure remains recognizable, without severe errors such as objects disappearing into thin air.
    \item \textbf{1-point:} Severe image distortion or blurriness, or instruments/tissues exhibiting teleportation, disappearance, or other illogical phenomena.
\end{itemize}

\subsection{Instrument Operation Plausibility}
\label{subsec:instrument-operation}

\textbf{Evaluation Focus:}  

Assessment of the plausibility of the instrument operation, including the instrument's appearance and physical handling. In particular, we evaluate the instrument's appearance (\textit{e.g.}, is it a real, non-hallucinated surgical instrument appropriate for the procedure?) and assess its action (\textit{i.e.}, whether the movement trajectories, manipulation techniques, and operational execution of surgical instruments are reasonable and technically sound). This includes appropriate instrument selection, movement paths, grasping angles, force application, and coordination between instruments.

\textbf{Scoring Criteria:}
\begin{itemize}

    \item \textbf{5-point:} The instruments shown are correct, realistic, and visually indistinguishable from a real surgical tool. Instrument movements are precise and technically proficient, mirroring expert surgical technique.
    
    \item \textbf{4-point:} The instruments shown are accurate and reasonable.  Instrument movements have minor technical imperfections (\textit{e.g.}, slightly suboptimal angles, minor inefficiencies, or less fluid motion) that remain effective.
    
    \item \textbf{3-point:} The instruments shown are recognizable but have minor flaws or locally unrealistic details (\textit{e.g.}, slightly incorrect proportions or texture). Instrument movements are in generally correct directions but are executed clumsily or inefficiently, such as awkward grasping angles, hesitant movements, or suboptimal instrument positioning that reduces operational effectiveness.
    
    \item \textbf{2-point:} The instruments shown are not very realistic (\textit{e.g.}, have distorted shape or clearly incorrect features) but are still loosely recognizable as a surgical tool. Instrument movements exhibit obvious technical problems, such as inappropriate trajectories or ineffective manipulations.
    
    \item \textbf{1-point:} The instruments shown have serious violations. This includes the generation of fabricated, non-existent, or physically impossible instruments, or real instruments moving in physically impossible ways or performing actions they are not designed for (\textit{e.g.}, attempting to cut with grasping forceps).

\end{itemize}

\subsection{Environment Feedback Plausibility}
\label{subsec:environment-feedback}

\textbf{Evaluation Focus:} Assessment of whether the feedback of the surgical scene (tissues, organs, etc.) to surgical operations is realistic and conforms to physical laws and anatomical knowledge. This dimension specifically measures whether the direct consequences of instrument actions (\textit{e.g.}, tissue deformation, bleeding patterns, eschar formation) are physically and anatomically accurate when compared to the reference video.

\textbf{Scoring Criteria:}
\begin{itemize}
    \item \textbf{5-point:} The feedback completely conforms to reality, such as morphological changes after traction, bleeding patterns after cutting, eschar formation after coagulation, etc., all appearing highly realistic with correct anatomical structures.
    \item \textbf{4-point:} The feedback is generally realistic and credible, with physical and biological responses largely correct, but with minor inaccuracies in details, such as slight deviations in the magnitude of tissue deformation, somewhat unnatural bleeding volume or diffusion speed, or imprecise degree of tissue discoloration after coagulation. These deviations do not affect the overall judgment of surgical scene authenticity.
    \item \textbf{3-point:} The feedback is partially realistic but with obvious deviations, such as incorrect bleeding volume or color, or tissue deformation appearing somewhat rigid.
    \item \textbf{2-point:} The feedback exhibits significant problems, such as minimal or severely incorrect bleeding patterns from cut tissues, obviously unnatural movement patterns when tissues are pulled, or noticeable errors in anatomical structures (\textit{e.g.}, incorrect organ position relationships). These issues are sufficient to raise concerns among experienced surgeons but have not reached the level of completely violating medical common sense.
    \item \textbf{1-point:} Serious violations of physical or medical common sense occur. For example: no bleeding after cutting a major blood vessel; after removing an organ, the exposed anatomical structures below are completely incorrect; tissues are stretched by instruments to impossible lengths without tearing.
\end{itemize}

\subsection{Surgical Intent Plausibility}
\label{subsec:surgical-intent}

\textbf{Evaluation Focus:} Assessment of whether the predicted surgical operations, as displayed from the input frame up to the current time point (\textit{e.g.}, 1-second, 3-second, or 8-second), demonstrate appropriate surgical intent and strategic reasoning for the current procedure. This evaluates whether the overall purpose, goal, and cumulative surgical effect of the predicted actions are consistent with logical surgical progression, appropriate procedural steps, and sound clinical decision-making when compared to the reference video.

\textbf{Scoring Criteria:}
\begin{itemize}
    \item \textbf{5-point:} The surgical intent is completely appropriate for the current procedure, with operations demonstrating a clear, logical purpose that aligns perfectly with standard surgical protocols and clinical objectives. The predicted actions show coherent strategic planning.
    \item \textbf{4-point:} The surgical intent is generally appropriate and clinically reasonable, with operations aligned with the current procedure, but the strategic choices are slightly suboptimal or less efficient than ideal approaches. The underlying surgical reasoning remains sound despite minor strategic inefficiencies.
    \item \textbf{3-point:} The surgical intent is discernible and partially appropriate, but shows questionable alignment with the current procedure. Operations suggest unclear priorities or strategic confusion, though some logical connection to surgical goals can still be identified.
    \item \textbf{2-point:} The surgical intent is unclear or obviously mismatched with the current procedure. Operations appear to lack coherent purpose, address already-completed objectives, or demonstrate significant strategic errors. However, some basic surgical reasoning can still be inferred, even if fundamentally flawed.
    \item \textbf{1-point:} Complete absence of appropriate surgical intent or serious violations of surgical logic. For example: performing tissue dissection in areas where dissection is already complete and hemostasis achieved; attempting coagulation when the priority should be addressing active bleeding; or conducting operations that contradict the evident clinical needs of the current procedure.
\end{itemize}

\section{Qualitative Case Studies of Generated Videos}\label{sec:good-bad-examples}

To provide deeper insight into the model performance beyond aggregated scores, this section presents a case-by-case analysis of representative high-scoring (Fig.~\ref{fig:good-example}) and low-scoring (Fig.~\ref{fig:bad-example}) examples from the SurgVeo benchmark, contextualized with synthesized expert feedback.

\subsection{Analysis of High-Scoring Examples}
\label{appendix:high_scoring}

\begin{figure*}[t]
	\begin{center}
		\includegraphics[width=0.9\linewidth]{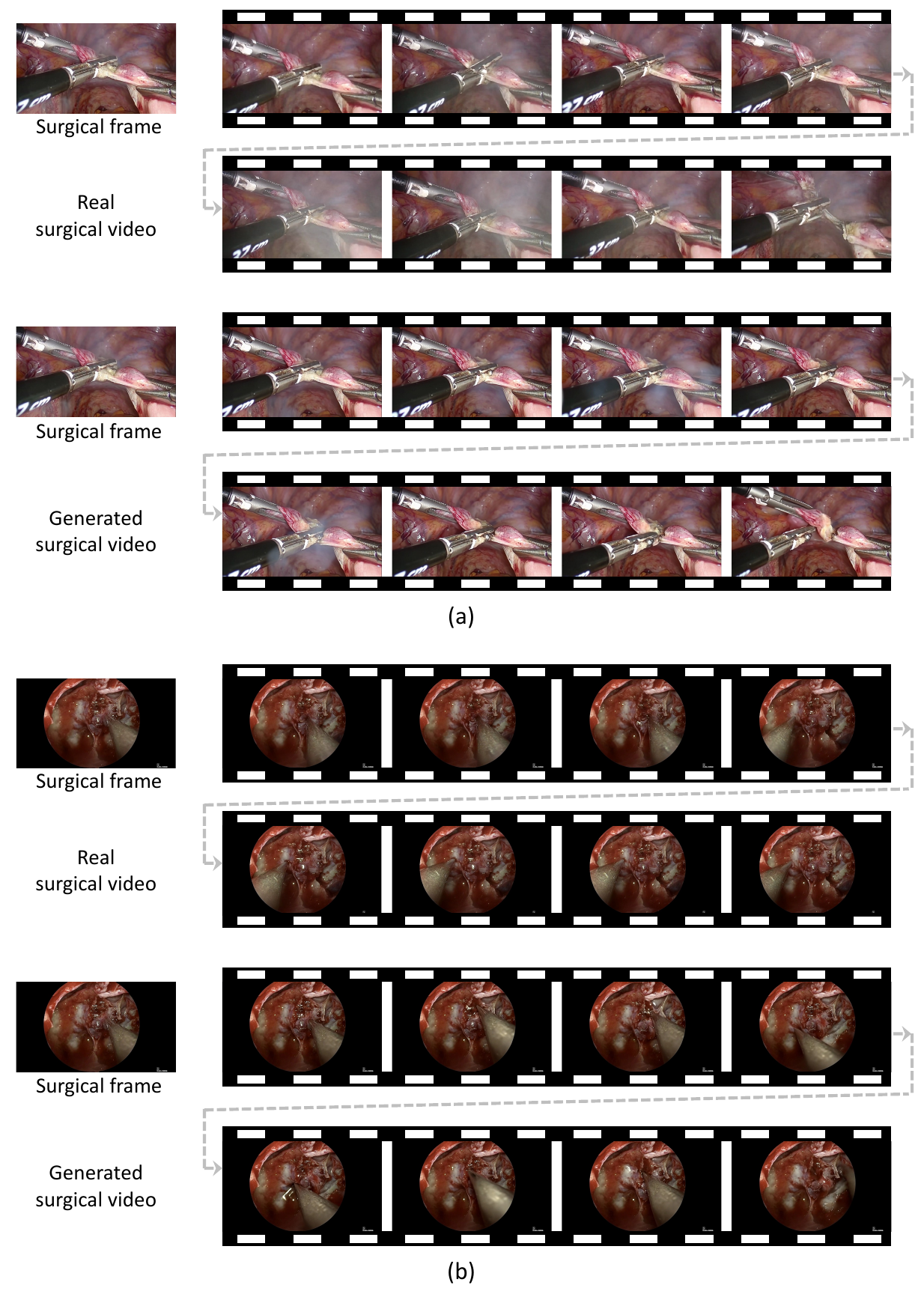}
	\end{center}
	\caption{Qualitative examples of high-scoring video generations from the SurgVeo benchmark. Each pair shows the real surgical video (top row) and the generated video (bottom row) evolving from the same starting frame. (a) A laparoscopic procedure where the Veo model generates a naturally flowing and plausible dissection. (b) A neurosurgical procedure where the generated video is nearly identical to the real reference video, presenting a high-quality case.}
	\label{fig:good-example}
\end{figure*}

The examples in Fig.~\ref{fig:good-example} illustrate the upper bound of the Veo model's current capabilities, showcasing success primarily in mimicking visually plausible, often simple or even non-operative, actions.

\begin{itemize}
    \item \textbf{Laparoscopic Dissection:} Experts consider this a strong generation, noting that the dissection appears "relatively natural" with "few errors" in Fig.~\ref{fig:good-example}(a). This suggests the Veo model can effectively replicate continuous, visually predictable actions like smooth tissue manipulation when the underlying surgical logic is straightforward.

    \item \textbf{Neurosurgical Pause:} This case highlights a high-quality generation in Fig.~\ref{fig:good-example}(b). The generated video is visually "almost identical" to the reference, where the Veo model successfully mirrors this calm period.
\end{itemize}

\subsection{Analysis of Low-Scoring Examples}
\label{appendix:low_scoring}

\begin{figure*}[t]
	\begin{center}
		\includegraphics[width=0.9\linewidth]{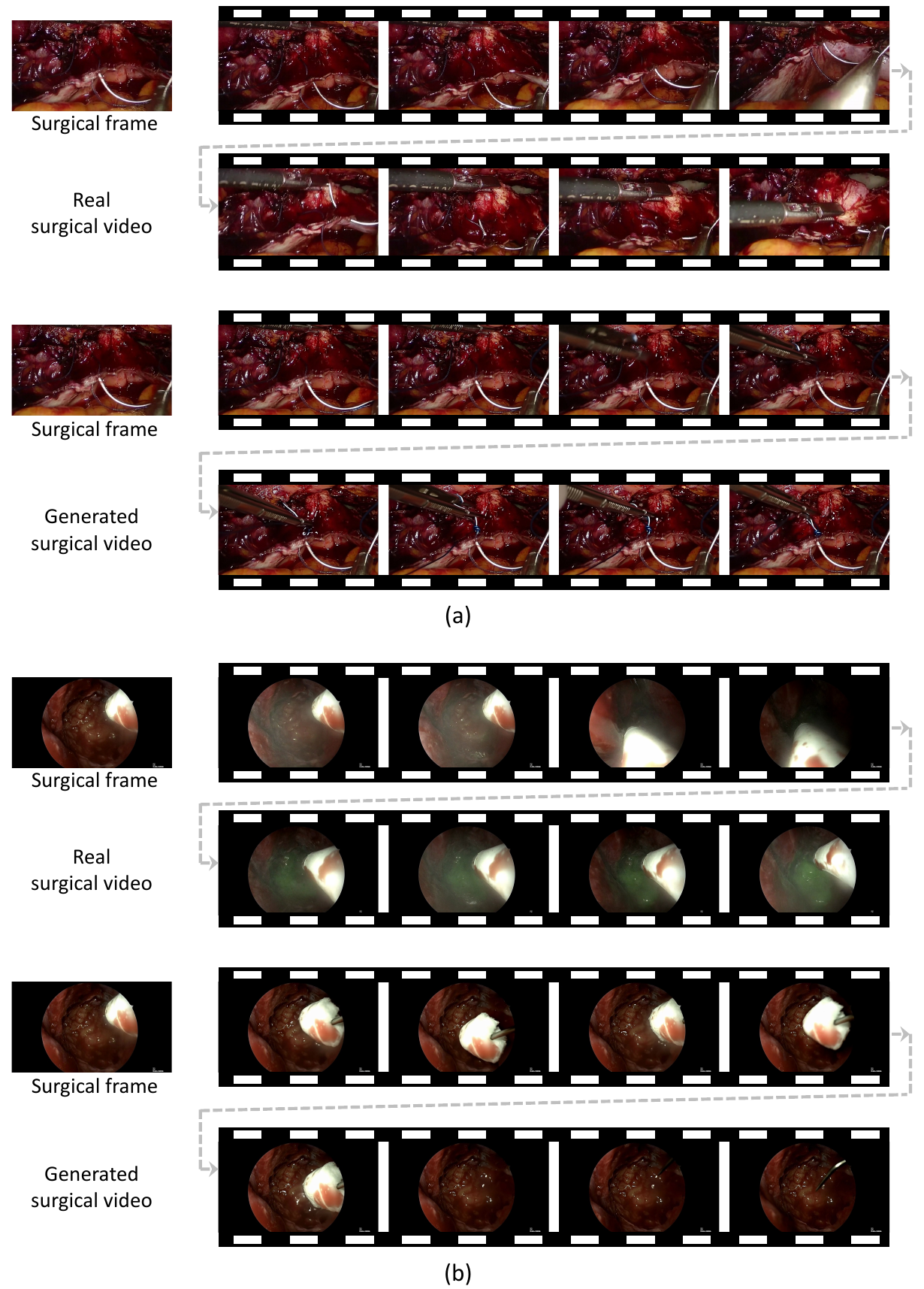}
	\end{center}
	\caption{Qualitative examples of low-scoring video generations, demonstrating catastrophic failures in plausibility. Each pair shows the real surgical video (top row) and the generated video (bottom row). (a) A laparoscopic procedure where the Veo model is expected to perform suturing. Instead, it hallucinates a fabricated surgical instrument and performs an unrecognizable, non-standard operation. (b) A neurosurgical procedure where the real intent is to apply biologic glue. The generated video completely misses this intent, failing on all three high-level SPP dimensions despite acceptable visual quality.}
	\label{fig:bad-example}
\end{figure*}

The examples in Fig.~\ref{fig:bad-example} demonstrate the catastrophic failures in surgical logic identified quantitatively in Section \ref{section_video_error_analysis}, often occurring despite acceptable visual quality.

\begin{itemize}
    \item \textbf{Laparoscopic Suturing Attempt:} Experts identify failures across multiple SPP dimensions, as illustrated in Fig.~\ref{fig:bad-example}(a). The Veo model completely fails to recognize the intended complex action (suturing). It hallucinates a "non-existent" surgical instrument and depicts an "unrecognizable, non-standard operation". Consequently, the simulated "tissue feedback is also poor". This case exemplifies a cascade of errors stemming from a failure to understand both the required action and the appropriate tools.

    \item \textbf{Neurosurgical Glue Application:} This example represents a complete failure at the higher levels of the SPP in Fig.~\ref{fig:bad-example}(b), despite adequate visual rendering. Experts state that while the "visual quality itself is acceptable, everything else, instrument generation, surgical intent, operation, and feedback, is wrong". The Veo model entirely misses the crucial surgical intent of applying biologic glue, instead generating unrelated and nonsensical actions. This starkly illustrates the Veo model's inability to grasp the strategic purpose (apex of the SPP) and translate it into a coherent sequence of plausible actions and consequences.
\end{itemize}

\section{Public Release and Reproducibility}
\label{sec:public-release-reproducibility}

To foster further research and enable rigorous comparison of future models, the complete SurgVeo benchmark will be publicly released at \url{https://github.com/franciszchen/SurgVeo}. The release includes:

\begin{itemize}
    \item All input frames and real surgical continuation videos
    \item Generated video outputs from Veo-3 for each benchmark sample
    \item Anonymized expert evaluation scores across all four assessment dimensions
    \item Detailed evaluation protocols and scoring guidelines
\end{itemize}

This comprehensive release enables the research community to reproduce our findings, benchmark alternative generative models, and advance the understanding of world modeling capabilities in specialized professional domains.

\end{document}